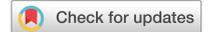

# OPEN  Automated analysis of fibrous cap in intravascular optical coherence tomography images of coronary arteries

Juhwan Lee[1], Gabriel T. R. Pereira[2], Yazan Gharaibeh[3], Chaitanya Kolluru[1], Vladislav N. Zimin[2], Luis A. P. Dallan[2], Justin N. Kim[1], Ammar Hoori[1], Sadeer G. Al-Kindi[2], Giulio Guagliumi[4], Hiram G. Bezerra[5] & David L. Wilson[1,6 ✉]

Thin-cap fibroatheroma (TCFA) and plaque rupture have been recognized as the most frequent risk factor for thrombosis and acute coronary syndrome. Intravascular optical coherence tomography (IVOCT) can identify TCFA and assess cap thickness, which provides an opportunity to assess plaque vulnerability. We developed an automated method that can detect lipidous plaque and assess fibrous cap thickness in IVOCT images. This study analyzed a total of 4360 IVOCT image frames of 77 lesions among 41 patients. Expert cardiologists manually labeled lipidous plaque based on established criteria. To improve segmentation performance, preprocessing included lumen segmentation, pixel-shifting, and noise filtering on the raw polar ($r, \vartheta$) IVOCT images. We used the DeepLab-v3 plus deep learning model to classify lipidous plaque pixels. After lipid detection, we automatically detected the outer border of the fibrous cap using a special dynamic programming algorithm and assessed the cap thickness. Our method provided excellent discriminability of lipid plaque with a sensitivity of 85.8% and A-line Dice coefficient of 0.837. By comparing lipid angle measurements between two analysts following editing of our automated software, we found good agreement by Bland–Altman analysis (difference 6.7° ± 17°; mean ~ 196°). Our method accurately detected the fibrous cap from the detected lipid plaque. Automated analysis required a significant modification for only 5.5% frames. Furthermore, our method showed a good agreement of fibrous cap thickness between two analysts with Bland–Altman analysis (4.2 ± 14.6 µm; mean ~ 175 µm), indicating little bias between users and good reproducibility of the measurement. We developed a fully automated method for fibrous cap quantification in IVOCT images, resulting in good agreement with determinations by analysts. The method has great potential to enable highly automated, repeatable, and comprehensive evaluations of TCFAs.

Thin-cap fibroatheroma (TCFA) and plaque rupture have been recognized as the most frequent risk factor for thrombosis and acute coronary syndrome (ACS)[1,2]. Based on histologic studies, a fibrous cap thickness of < 65 µm is implicated in plaque ruptures[3]. However, Kume et al. found thicker cap measurements with intravascular optical coherence tomography (IVOCT), likely due to tissue shrinkage in histology[4]. The IVOCT consensus document also suggested that this cutoff should be adjusted when applied to IVOCT images to account for the potential tissue shrinkage (10–20%) during histopathologic processing[5]. Point measurements of thickness are inadequate because biomechanics indicates that the both size and distribution of the thinned region is important for rupture[6]. This biomechanical perspective suggests the need for a three-dimensional (3D) analysis of cap thickness. Although intravascular ultrasound is widely used to visualize coronary arteries, it is not possible to identify the presence of TCFA due to its low resolution (150–250 µm)[7,8]. As a high contrast, high-resolution imaging technique, IVOCT provides an axial resolution of approximately 12–18 µm[9]. With its near histologic resolution and its optical contrast, IVOCT allows a unique assessment of plaque microscopic features (e.g., TCFA, macrophage, cholesterol crystal, and microchannel), and in particular cap thickness, suggesting an improved opportunity to assess the plaque vulnerability.

[1]Department of Biomedical Engineering, Case Western Reserve University, Cleveland, OH 44106, USA. [2]Harrington Heart and Vascular Institute, University Hospitals Cleveland Medical Center, Cleveland, OH 44106, USA. [3]Department of Biomedical Engineering, The Hashemite University, Zarqa 13133, Jordan. [4]Cardiovascular Department, Galeazzi San'Ambrogio Hospital, Innovation District, Milan, Italy. [5]Interventional Cardiology Center, Heart and Vascular Institute, University of South Florida, Tampa, FL 33606, USA. [6]Department of Radiology, Case Western Reserve University, Cleveland, OH 44106, USA. ✉email: dlw@case.edu

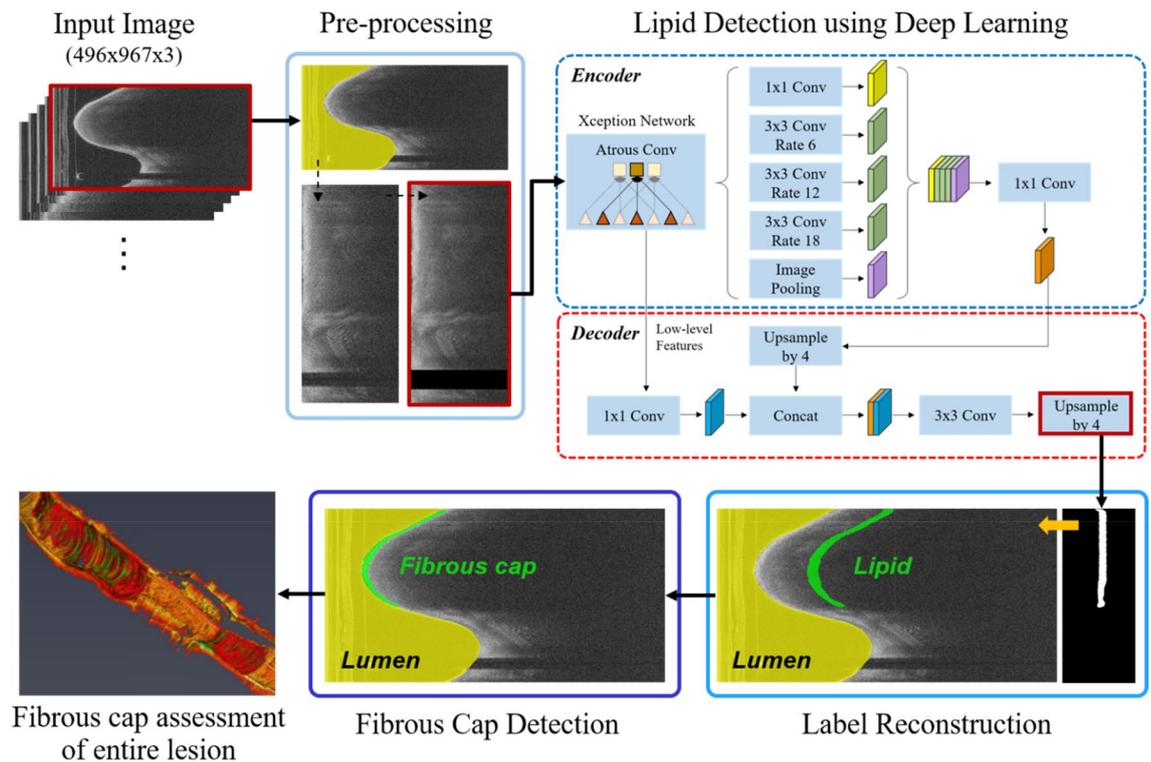

**Figure 1.** An overall workflow of the proposed method. Preprocessing including lumen segmentation, pixel-shifting, and noise reduction is applied to the original IVOCT ($r, \theta$) image, and the output is used as an input to DeepLab-v3 plus deep learning model for lipid plaque detection. In the context of detected lipid plaque, fibrous cap is detected and quantitatively assessed.

Although IVOCT enables clinicians to identify the presence of TCFA, challenges remain. First, a single IVOCT pullback includes 300–500 image frames, leading to data overload. Typically, a clinician needs 1 hour to carefully review a pullback and label lipidous plaque and fibrous cap thickness. This timeframe is impossible during a clinical procedure and laborious for research studies with large numbers of pullbacks. Second, manual analysis of fibrous cap can be subject to high inter- and intra-observer variability[5]. This high degree of variability creates a confounding factor for widespread quantitative and visual evaluation, especially considering the variable experience among cardiologists. An accurate, automated method could provide faster, consistent assessment of TCFA and support treatment decision-making. Furthermore, quantitative assessment of the fibrous cap can support research studies to elucidate the underlying mechanisms and factors in cap thinning and plaque rupture. Compared with the results of variable manual assessments, automated, consistent measurements will improve the power to examine change in such studies.

The purpose of this study was to develop and test a fully automated and accurate approach to detect lipid plaque and quantitatively assess the fibrous cap in 3D IVOCT images. Our approach includes deep learning identification of lipidous plaques using a large, carefully labeled dataset, segmentation of the fibrous cap, numeric assessment of cap thickness, and evaluation of methods. For the first time, we will be able to automatically analyze cap thickness over an entire lesion, which will allow us to extend simple point measurements of minimum thickness to an assessment more closely realizing the full biomechanical impact of a lesion.

## Methods
Briefly, our method identifies the location of lipidous plaque using a semantic segmentation deep learning method with important preprocessing steps. Then, a specialized algorithm identifies the outer fibrous cap border and completes segmentation of the fibrous cap. The software performs multiple quantitative assessments (e.g., angle, thickness map, and thickness statistics). Next, we compare the results of this method with those of expert analysts. The following text describes the image data used for deep learning identification of lipidous plaques; the image processing and analysis methods, including deep learning; and our approaches to evaluation. Figure 1 shows an overall workflow of the proposed method on the representative IVOCT frame.

**Study population and manual labeling.** Images are from a sub-study of the TRiple Assessment of Neointima Stent FOrmation to Reabsorbable polyMer With Optical Coherence Tomography (TRANSFORM-OCT) trial[10]. Patients with stable angina and documented ischemia or ACS who had undergone IVOCT exami-nation were eligible for the study. Major exclusion criteria were the presence of unprotected left main disease, chronic total occlusion, baseline serum creatinine >2.0 mg/dL, life expectancy <18 months, and unsuitability to OCT imaging (at the investigator's discretion). Final analysis included a total of 4360 image frames of 77 lesions among 41 patients (43 pullbacks). Of the 4360 images, 2363 images were annotated as lipidic, and 1997 images were annotated as normal. The IVOCT images were acquired with a frequency-domain OCT system (ILUMIEN OPTIS, St. Jude Medical), which uses a tunable light source sweeping from 1250 to 1360 nm. Imaging pullback was performed with a frame rate of 180 fps, pullback speed of 36 mm/s, and axial resolution of approximately 20 µm. This study was conducted in accordance with the Declaration of Helsinki and with approval from the Institutional Review Board of University Hospitals Cleveland Medical Center (Cleveland, OH, USA). Written informed consent was waived by the Institutional Review Board of University Hospitals Cleveland Medical Center (Cleveland, OH, USA).

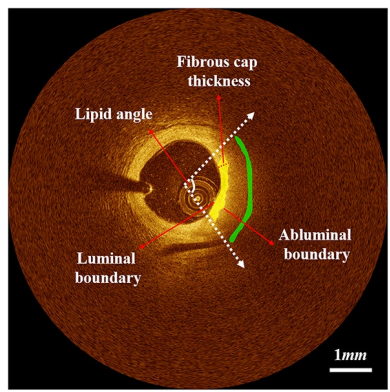

**Figure 2.** Representative case of lipid and fibrous cap annotations with quantitative measurements: lipid angle and fibrous cap thickness. Green and yellow are lipid and fibrous cap, respectively. The luminal boundary perfectly matches to the lumen boundary, and the abluminal boundary is defined as the inner border of the lipid pool. Lipid annotations were 50 pixels away from the lumen boundary with a ribbon of 20 pixels wide in $(r,\theta)$ domain. Colors are green (lipid plaque) and yellow (fibrous cap).

For manual annotation, the raw IVOCT data in the polar $(r, \theta)$ domain was transformed to the Cartesian $(x, y)$ domain. Each image was manually analyzed by two cardiologists from the Cardiovascular Imaging Core Laboratory. The location of lipid plaque was identified based on definitions in the "consensus document"[5] using OCTOPUS software[11]. Specifically, a signal-poor region with diffused outer borders, a fast IVOCT signal drop-off, and little or no IVOCT signal backscattering within a lesion was considered to be a lipid plaque. In case of disagreement between the two readers, they revaluated the frames and reached a consensus decision. Because lipid rapidly attenuates the IVOCT signal, it is impossible to see the outer edge of lipidous plaque in IVOCT images. As a result, lipid annotations for this study consisted of marking the arc of the lipidous region with a depth of 50 pixels from the lumen boundary and broad stroke in $(x, y)$ images. In our software, we ensured that every lipid label consisted of a ribbon of pixels 20 pixels wide in the $(r, \theta)$ representation. Every other pixel in the $(r, \theta)$ images was given a label of "other," which allowed us to set up an approach of binary semantic segmentation. An example of manual annotation is shown in Fig. 2.

**Preprocessing.** We applied a modified version of a previously proposed approach for preprocessing[12,13] to raw polar $(r, \theta)$ IVOCT data (Fig. 1). First, the lumen boundary was detected using the deep learning approach as described previously by our group[14]. Second, each A-line was pixel-shifted to the left so that all vessel wall A-lines have the same starting pixel along the radial direction. A-line shifting creates a smaller region of interest for deep learning and aligns tissues so that different lesions look more similar to the network. That is, in $(x, y)$, two lesions at different angles (e.g., 90° and 270°) and different depths look quite different to a network, but in our preprocessed $(r, \theta)$ representation, the two lesion may look quite similar, thereby reducing the number of training samples needed. Third, we cropped and used the first 1.5 mm (300 pixels) from the lumen boundary in the $r$ direction because of the limited penetration depth of the IVOCT signal, reducing the size of data in deep learning. Fourth, speckle noise was reduced using a Gaussian filter with a kernel using a standard deviation of 1 pixel and a footprint of (7,7). Preprocessed images were used as an input to the deep learning model for lipid plaque detection.

**Automated identification of lipid plaque.** We implemented DeepLab-v3 plus semantic segmentation[15] for lipid plaque segmentation as shown in Fig. 1. The DeepLab-v3 plus network takes advantage of both an encoder module that enables capture of contextual information at different grid scales and a decoder module that effectively recovers object boundaries. We briefly describe the three key concepts as follows. First, atrous convolution is a method that controls the resolution of features and adjusts the field-of-view of a filter. Atrous convolution allows the network to learn multiscale contextual information without increasing the number of parameters and generalizes standard convolution operation. Second, the depth-wise convolution carries out an independent spatial convolution for each input channel, and the pointwise convolution is then used to combine the output from the depth-wise convolution. This process drastically reduces computational complexity while maintaining similar—or better—segmentation performance. Third, the encoder extracts the essential information at different resolutions. The most important features are the objects in the image and their locations. Here we used the Xception[16] as the backbone network for feature extraction. After each convolution, the batch normalization and rectified linear unit were followed. The decoder generated an output label with the same image resolution as the input image. The receptive field is one of the key factors for determining segmentation performance because it provides context for a decision. In this study, the receptive field of our network covered the entire input image.

**Automated fibrous cap detection.** A fibroatheroma is characterized by a low IVOCT signal region with a poorly delineated border followed by a very rapidly attenuating signal in the lipidous region due to the associated high absorption of light[5]. In the context of detected lipid plaque, we assessed fibrous cap thickness between the luminal boundary and the abluminal boundary, defined as the boundary between the fibrous and lipidous regions[17] (see Fig. 2). The lumen boundary was obtained as previously described from the deep learning lumen segmentation. The abluminal boundary has a high-intensity difference between the fibrous cap and lipid pool, and therefore can be characterized by a gradual transition of pixel intensity from bright to dark. To detect the abluminal boundary, we used a dynamic programming as proposed previously by our group[17]. The method was previously validated against manual delineations in 14 lipid-rich lesions, and provided promising results[17]. Briefly, we generated a gradient map for each A-line using a filtering operation, which yielded a strong positive response at the bright to dark edges. Then, using dynamic programming, we found the path with the maximum cumulative edge strength, which defined the abluminal boundary. Details of these algorithms are provided elsewhere[17].

**Network training.** We used an adaptive moment estimation optimizer for network training[18]. The initial learning rate, drop factor, and drop period were empirically set to 0.001, 0.2, and 5, respectively. To gradually reduce the learning rate, we multiplied the initial learning rate by a drop factor for every drop period. The initial weights of each layer were determined using the Xception[16] pretrained on the ImageNet database[19]. Weights were fine-tuned starting from the last layer by changing the learning rates of the previous layers. Because our dataset was imbalanced, we computed the inversed median frequency of class proportions and used them as the class weight. To avoid overfitting during the training, we stopped the training when the validation loss did not improve over 5 consecutive epochs or when a maximum number of epochs (50) was reached, whichever occurred first. The binary cross-entropy loss function over the softmax function was extensively used as the output. All image processing and network training were done using MATLAB (R2018b, MathWorks) on a NVIDIA Geforce TITAN RTX GPU (24 GB memory).

**Performance evaluation.** We divided a total of 77 lipidic lesions into the training set and held-out test set. Using a 90–10% split, the training and held-out test set included 68 lesions (3821 frames) and 9 lesions (539 frames), respectively. To optimize network hyperparameters, we performed a five-fold cross-validation on the training dataset. The 68 lesions were split into 5 independent subsets. For each iteration, 3 were considered for training, 1 for validation, and 1 for testing. The held-out data were used for further evaluations.

Lipidous plaque segmentation was evaluated using conventional metrics, such as precision, sensitivity, specificity, and Dice coefficient. Particularly, given that A-line classification is more suitable for angle measurement of lipidous region and removes the confusion associated with the depth of unseen lipidous regions, we assessed results in terms of A-line classification[13,20]. In addition, we quantified the lipid angle and fibrous cap thickness for each frame (Fig. 2) and performed an inter-observer variability test between two analysts after automated analysis. Two experienced cardiologists independently edited the automated lipid plaque and fibrous cap detection results on the held-out test set using a dedicated editing tool. This approach provided two pairings of segmentation (lipid)/detection (fibrous cap) results. We evaluated the agreement using linear regression and Bland–Altman analyses. In addition, we asked two cardiologists to perform visual assessment of fibrous cap detection. They reviewed the fibrous cap detection results and responded to this question: "Is the automated fibrous cap detection acceptable?" Responses were scaled from 1 to 5 as follows: 1, strongly disagree; 2, disagree; 3, unsure; 4, agree; and 5, strongly agree.

## Results

**Patient samples.** This study included 41 patients with multi-vessel disease who had undergone staged percutaneous coronary intervention with stent implantation. No patients were excluded on the basis of clinical characteristics or image processing results. The median age was 63.7 ± 10.0 years, and 85.4% were men. Among the 41 patients, 28 (68.3%) had hypertension, 25 patients (61%) were current smokers, and 6 patients (14.6%) had diabetes mellitus. Table 1 details patient characteristics.

**Automated lipid detection.** Our method successfully identified the lipid plaque in almost all situations. Figure 3 shows the manual annotation and automated prediction on the held-out test set. Our method provided excellent discriminability of lipid plaque from other tissues such as fibrous tissue and calcification. On the held-out set, our method had a precision of 81.8%, sensitivity of 85.8%, specificity of 90.7%, and A-line Dice coefficient of 0.837, and the differences were within 2% compared with those of the five-fold cross-validation. These findings indicate that our model is suitably trained and reliable. The supplementary Figure S1 demonstrates training and validation learning curves for one cross-validation round, indicating a small performance differ-ence between training and validation. Inter-observer variability of lipid angle deviation between two analysts are provided in the supplementary Figure S2. A lipid angle difference between ground truth and automated prediction + manual editing is shown in the supplementary Figure S3.

| | |
|---|---|
| Age (years), mean ± SD | 63.7 ± 10.0 |
| Men, *n* (%) | 35 (85.4%) |
| Hypertension, *n* (%) | 28 (68.3%) |
| Hypercholesterolemia, *n* (%) | 17 (41.5%) |
| Current smoker, *n* (%) | 25 (61.0%) |
| Diabetes mellitus, *n* (%) | 6 (14.6%) |
| Body mass index (kg/m$^2$), mean ± SD | 41.4 ± 2.8 |
| Previous myocardial infarction, *n* (%) | 5 |
| Previous PCI, *n* (%) | 5 |
| Previous CABG, *n* (%) | 1 |
| Previous stroke, *n* (%) | 4 |
| Indication for PCI, *n* (%) | |
| ACS | 8 |
| Stable angina | 9 |
| Unstable angina | 4 |
| NSTEMI | 5 |
| STEMI | 15 |

**Table 1.** Baseline characteristics of study cohort (*n* = 41). *PCI* percutaneous coronary intervention, *CABG* coronary artery bypass graft, *ACS* acute coronary syndrome, *NSTEMI* non-ST segment elevation myocardial infarction, *STEMI* ST-segment elevation myocardial infarction.

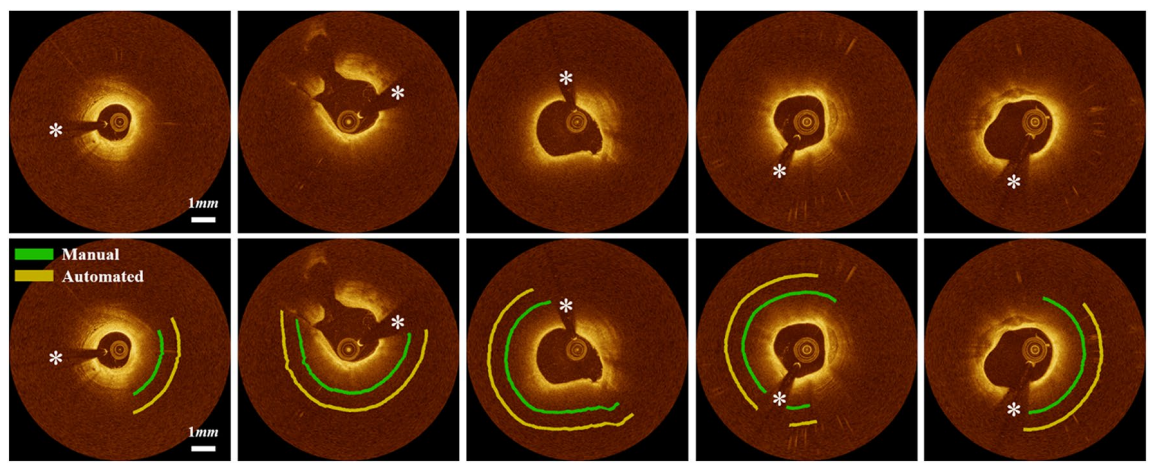

**Figure 3.** Lipid detection results on the held-out test set. Panels are (top) original IVOCT images and (bottom) manual and automated results. Green is the manual annotation, and yellow is the predicted result. *Please understand that these markups show the angle region of lipidic lesions.* The white asterisk (*) indicates the guidewire shadow.

Several cases challenged our method. Figure 4 shows the representative challenging cases that required modifications for detecting lipidous plaque. When the lipid was distributed near the side branch, our method generated unreliable predictions (Fig. 4A); however, these instances were rare. In addition, our method occasionally produced false predictions for the mixed plaques (Fig. 4B).

In addition, we evaluated the inter-observer variability of lipid angle measurement after manual editing between two experts (Fig. 5). Bland–Altman analysis showed a very small bias (6.7° ± 17.3°), and most of these measurements were included within the limit of agreement. Linear regression analysis also provided a very high similarity between two observers ($R^2$ = 0.943), indicating the exceptionally low user bias of the proposed method.

**Automated fibrous cap detection.** Our method accurately detects the fibrous cap from the detected lipid tissue. Figure 6 demonstrates automated fibrous cap detections on the representative cases. Our method required a significant modification for only 5.5% ± 0.9% frames of the held-out test set; of the 539 frames, Expert 1 modified 33 and Expert 2 modified 26. In these cases, most errors resulted from the false automated detec-tion of lipid. Furthermore, the two cardiologists unanimously scored with the highest value (5) for all lesions, indicating strong agreement that clinical decision-making would be the same for manual and automated results.

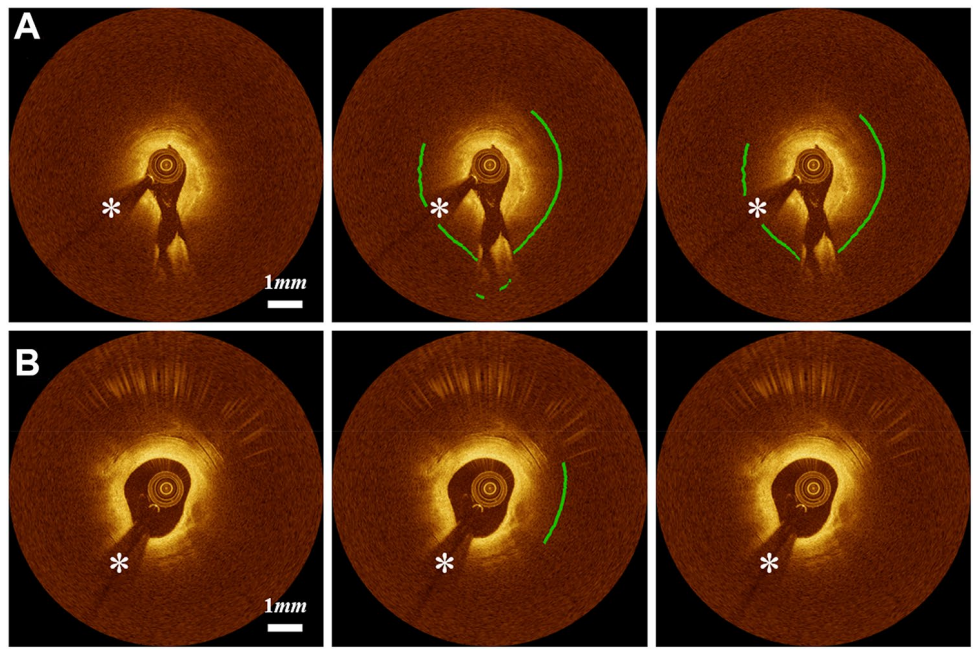

**Figure 4.** Challenging cases in lipid detection requiring editing. (**A**) The algorithm reasonably identified the lipid location, but showed errors on the side branch regions. (**B**) False lipid prediction often happened to the mixed plaques. The first, second, and third columns are original images, automated results, and manually edited results, respectively. Green is the fibrous cap, and white asterisk (\*) indicates the guidewire shadow.

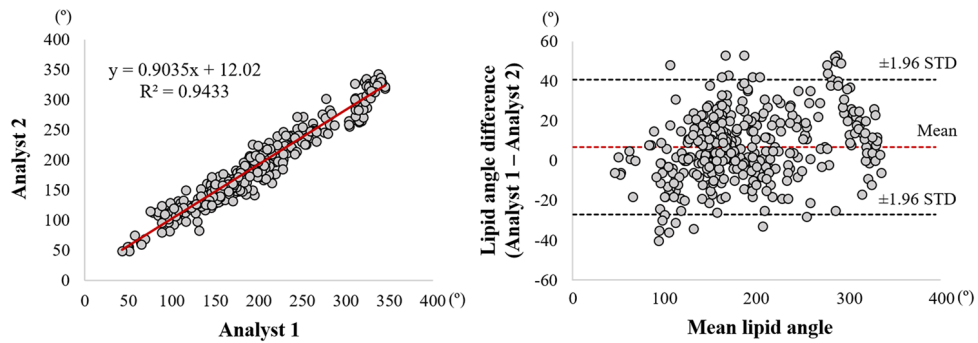

**Figure 5.** Inter-observer variability test of lipid angle between two analysts after automated prediction + manual editing. Panels are (left) linear regression plot and (right) Bland–Altman plot. The R-squared ($R^2$) value was 0.943 (left), and the mean bias between two analysts was 6.7° ± 17.3° (right).

Although our method worked well in most cases, several challenging cases required significant manual corrections. Our method produced unreliable detection results when the lipid label included other tissue types, such as tissue dissection (Fig. 7).

Bland–Altman analyses of fibrous cap thickness after editing showed a great reliability of the proposed method (Fig. 8). The bias of fibrous cap thickness measurement between two experts was 4.2 ± 14.6 μm, and these measurements were all included in the limit of agreement. From the linear regression analysis, the $R^2$ value of fibrous cap thickness measurement was 0.974, indicating no significant bias between two experts for using the software (Fig. 8). Figure 9 shows 3D visualizations of fibrous cap thickness on the representative IVOCT pullbacks with a short lesion with TCFA (Fig. 9A), a long lesion with TCFA (Fig. 9B), a short lesion without TCFA (Fig. 9C), and a long lesion without TCFA (Fig. 9D). Our method enables comprehensive assessment of fibrous cap thickness in the entire pullback.

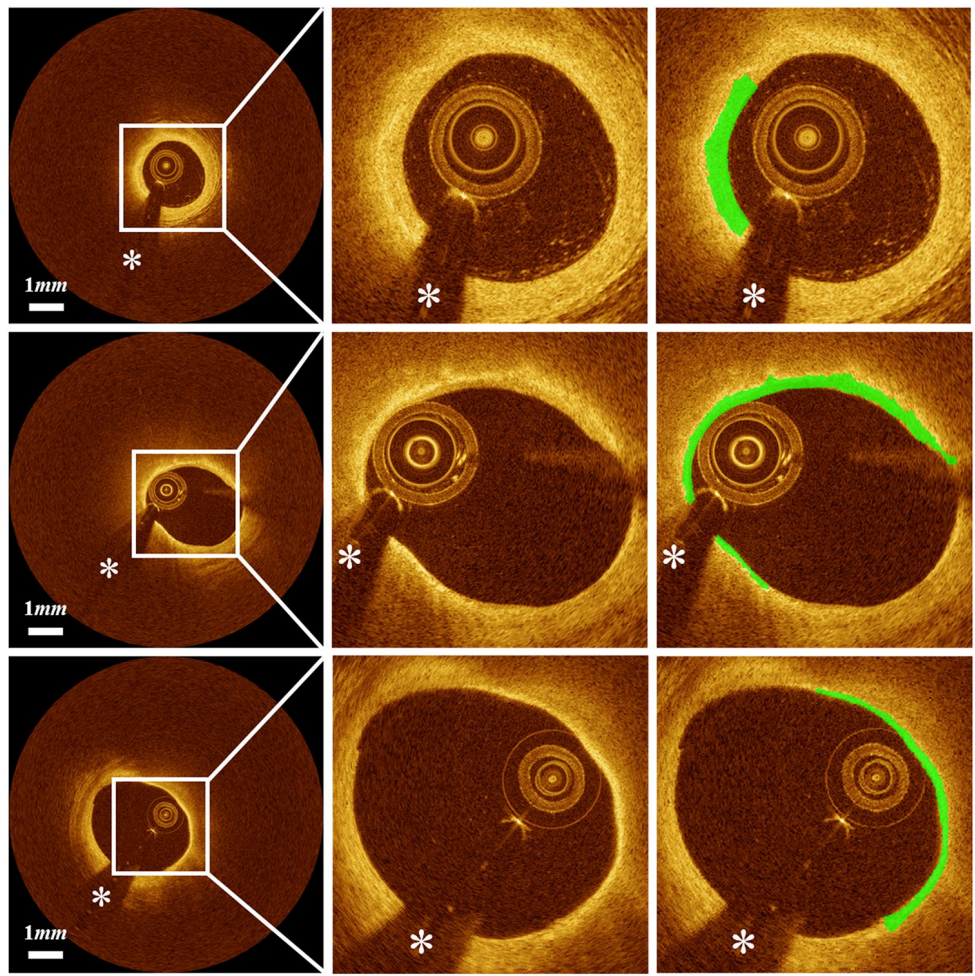

**Figure 6.** Automated thin-cap fibroatheroma detection results on the held-out test set: (left) original IVOCT image, (center) cropped image for better visualization, and (right) cropped image overlaid with fibrous cap detection. Green is fibrous cap, and white asterisk (*) is the guidewire shadow.

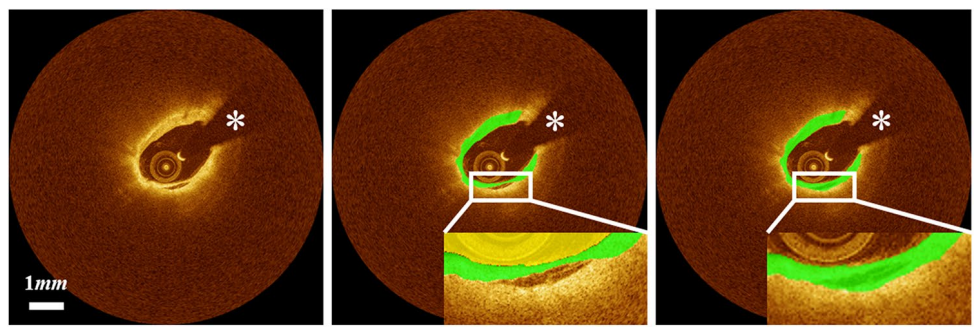

**Figure 7.** Challenging cases in fibrous cap detection requiring editing. Fibrous cap detection was not reliable when the lipid label includes other regions such as tissue dissection. The left, center, and right panels are original image, automated result, and manually edited results, respectively. Green is the fibrous cap, and white asterisk (*) indicates the guidewire shadow.

## Discussion

It is clinically accepted that the presence of TCFA is associated with plaque ruptures[1,2]. Although manual analysis of the fibrous cap is challenging and prone to high inter- and intra-observer variability, no studies have attempted to develop an automated image analysis method for the fibrous cap. We built on our previous studies[12–14,17,20–24] and for the first time have developed a fully automated method for quantitatively assessing the fibrous cap in

www.nature.com/scientificreports/

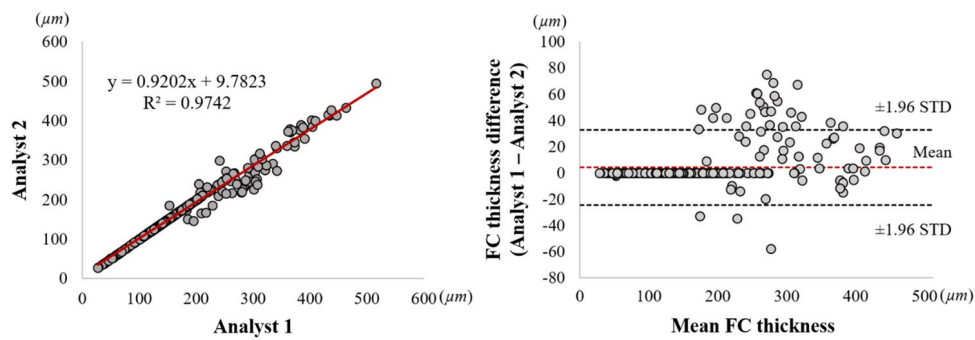

**Figure 8.** Inter-observer variability test of minimum fibrous cap thickness between two analysts after automated detection + manual editing. Panels are: (left) linear regression plot and (right) Bland–Altman plot. The R-squared ($R^2$) value between two analysts was 0.974 (left), and the mean bias was 4.2 ± 14.6 μm (right). This result indicates that there is no significant bias between two experts for using software.

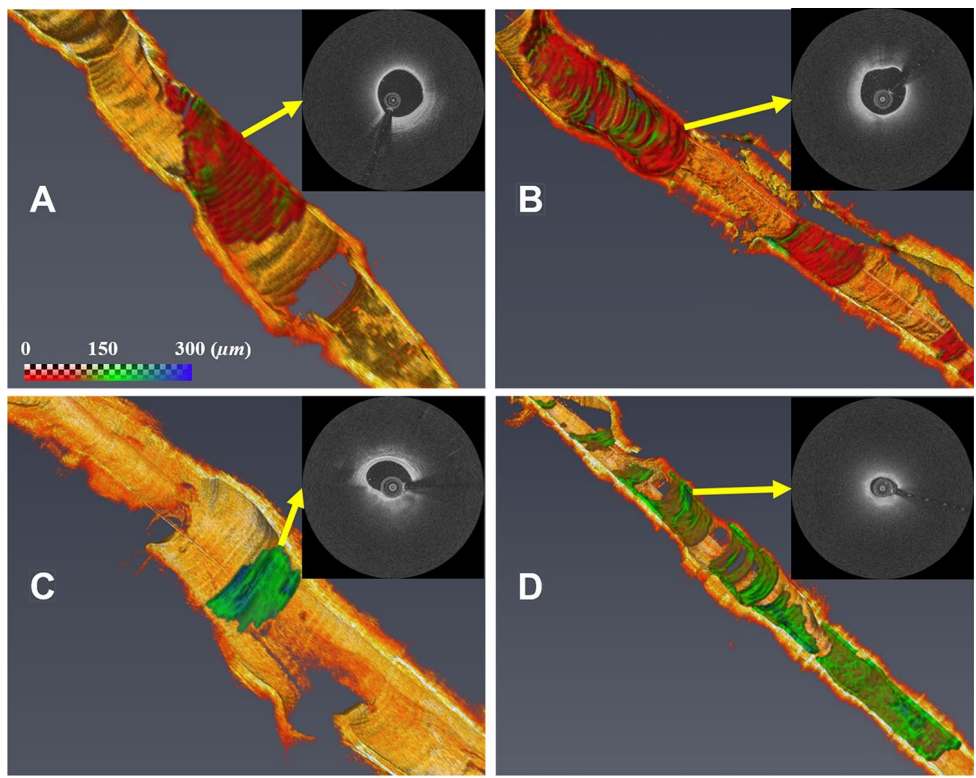

**Figure 9.** Three-dimensional (3D) visualizations of fibrous cap thickness on the representative IVOCT pullbacks, including: (**A**) short lesion with TCFA, (**B**) long lesion with TCFA, (**C**) short lesion without TCFA, and (**D**) long lesion without TCFA. The reader can zoom in each artery to see variations of fibrous cap thickness. (**A**) Although the lesion length was not too long (< 7 mm), the average fibrous cap thickness was less than 65 μm across the lesion indicating that the lesion is prone to rupture. (**B**) There were two lipidous lesions having 15 mm (left) and 5 mm (right) lengths. Both lesions were heavily lipidic with a mean cap thickness of < 65 μm. The artery was much more prone to rupture than (**A**). (**C**) The lesion was stable, since the length was short (< 3 mm) and the fibrous cap thickness was always greater than 150 μm. (**D**) Although the fibrous cap thickness was always over 80 μm across the lesion, the lesion length was very long (> 30 mm). There were several spots approaching toward the vulnerable plaque than (**C**). The color map visualizes the fibrous cap in the range of 0 to 300 μm. The yellow arrows indicate representative IVOCT frames of each rendering. Our method provides comprehensive fibrous cap map in the entire IVOCT pullback, so clinicians can make appropriate treatment decisions.

IVOCT images. Although we proposed a method for identifying fibrous caps in IVOCT images in our previous study[17], there were two significant limitations. First, the lipidous plaque must be manually determined by the user. Second, the method was validated on only 14 lipidous lesions (323 frames). In combination with automated lipid plaque segmentation using deep learning, our method provides very good results for fibrous cap detection with a reasonable computation time, suggesting that this method could be a promising solution for both research and clinical applications.

Despite the complexity of lipidous plaque segmentation, our method provides excellent agreement with experienced analysts. Results were very close to manual annotations, with a sensitivity of 85.8%, a specificity of 90.7%, and A-line Dice of 0.837. Accurate detection of lipidous plaque is a prerequisite for automated fibrous cap analysis. Although a good inter-reader agreement was observed between our expert analysts (Fig. 5), it is important to note that manual analysis can depend on the reader's experience. That is, if less experienced analysts are reading the images, results are subject to a high inter-observer variability compared with other tissues visualized with IVOCT, such as calcification. To minimize the potential discrepancy from manual annotation, analysts must very carefully annotate lipid tissues according to the definitions described previously[5]. Principally, if the signal quickly decays with a diffuse outer border, this region was regarded as lipid plaque. Based on these consistent labels, our deep learning model was able to successfully learn the lipid characteristics and clearly differentiate them from other plaques, such as fibrous and calcified plaques. Interestingly, most errors occurred in areas of mixed tissues (Fig. 4B), which included a certain level of lipid, and not from the fibrous or calcified plaques. This finding indicates that our method has the potential to detect mixed plaque, which is a forerunner of vulnerable lipidous plaque.

In addition to the lipid segmentation, our automated method showed a promising result for fibrous cap detection. The visual assessment of the fibrous cap border is challenging because of its thin appearance, giving rise to inaccurate measurement. Particularly, although IVOCT has a high axial/radial resolution, it is very difficult to manually assess the thinnest fibrous caps in IVOCT images. Given that the abluminal boundary of the fibrous cap shows a high-intensity difference between the fibrous cap and lipid pool, our boundary optimization method was suitable for capturing the gradual transition of pixel from bright to dark. Also, this method does not require any training process, thereby providing reproducible, consistent results.

Our method enabled excellent repeatability of the lipid and fibrous cap analyses by two expert analysts. From the inter-observer variability test, we found a very small bias (lipid angle, 6.7°±17.3°; fibrous cap thickness, 4.2±14.6 μm) and high similarity ($R^2$, lipid angle, 0.943; fibrous cap thickness, 0.974) between the two analysts. Our highly automated analysis will be exactly repeatable, and even with manual editing, it will likely be much more repeatable than using different analysts. For the fibrous cap, the two experts unanimously provided the highest visual assessment score (5) across the held-out test set (9 lesions with 539 frames) indicating that our method would lead to the same clinical decision-making as the manual assessment in clinics. In addition, automated analysis required a user intervention only for 5.5% frames on the held-out test set. Typically, the intervention consisted of changing the angle to be analyzed, a quick change. To review and edit automated analysis requires about 4 min on the typical vessel. Likely this time will improve with more experience and possibly improved editing tools. We estimate that a fully manual 3D analysis of fibrous cap thickness would take an analyst about 45 min on that same vessel or 11 times that of the semi-automated approach. This estimation suggests that our method will enable large research studies. In addition, with further refinement, it is conceivable that the method could be used clinically to help physicians determine a proper landing zone for a stent.

Automated 3D fibrous cap assessment, such as for cap size and distribution, may enhance clinical research as seen in the following examples. First, we can uniquely analyze serial imaging data on the long-term characterization (18 months) of stented vessels with the evolution of neo-atherosclerosis[10]. This approach will allow to determine characteristics of atherosclerosis and suboptimal stent deployment predictive of short- and intermediate-term adverse results—in-stent restenosis and plaque progression. Second, there is a unique opportunity to identify determinants of a future major adverse cardiovascular event (MACE). Risk prediction for MACE has met with some success using the computed tomography (CT) calcium score[25–28], CT angiography[29–32], and, more recently, with intravascular ultrasound[33]. However, none of these methods show the disease detail, such as fibrous cap thickness, size, and distribution that is available with IVOCT, suggesting a new opportunity to learn about the characteristics of atherosclerosis that are most detrimental to human health. We can analyze the characteristics of pre-stent atherosclerosis and stent deployment, as well as determine the risk of MACE using machine/deep learning approaches. Third, the analysis of lipidous, vulnerable plaques will suggest more aggressive drug therapeutic regimens. Guagliumi et al. showed that high-intensity lipid-lowering therapy can cause regression of those plaques[10], making the plaque more stable and less prone to rupture. With the use of our method, a clinician would be able to precisely segment the lipid and fibrous cap, ultimately assessing the effect of those medications on them.

Our method could enable a comprehensive assessment of plaque vulnerability. Macrophage, cholesterol crystal, and microchannel are commonly considered to be among the most important clinical predictors for plaque vulnerability in IVOCT images[34]. According to the previous report[35], an isolated few patterns are probably not clinically significant, whereas a large accumulation of these characteristics within a thin fibrous cap may be of more concern. In addition, Nakazato et al. reported that a higher prevalence of adverse plaque characteristics can be observed when the macrophage coexists with the TCFA in the context of fibrous cap[36]. In our previous study[37–39], we proposed automated methods for quantifying macrophage and microchannel in IVOCT images. Therefore, combining 3D assessments of fibrous cap thickness with the presence of macrophages, cholesterol crystals, and microchannels could aid a more comprehensive assessment of coronary plaque in IVOCT images.

Our method is computationally realistic for clinical and research applications. On our computer system with non-optimized code, the computation time is only about 0.1 s per image: 0.05 s for preprocessing, 0.02 s for lipid segmentation, and 0.03 s for fibrous cap detection. The proposed method can process a high-resolution IVOCT pullback of 540 frames within 1 min, which is suitable even for clinical application. Moreover, code optimization in a language such as C++ may allow our method to be suitable for real-time treatment planning.

This study has several limitations. First, our method often showed inaccurate lipid segmentations on the side branch and mixed plaque. The false detection arising at side branches is easily recognized and manually corrected. In addition, it may be possible to create an algorithmic fix. Regarding the mixed plaque, no widely accepted standard in IVOCT images exists because it significantly depends on the user's experience. However, the mixed plaques can also be easily identified and manually edited. Second, although we used a large dataset, a future study using even more data with accurate annotation may further improve performance. Third, we used a conventional deep learning semantic segmentation (DeepLab-v3 plus) for lipid segmentation. Results may be improved with the use of more advanced deep learning models.

## Conclusion

We developed a fully automated method for fibrous cap quantification in IVOCT images. Deep learning semantic segmentation well identified the location of lipid plaque, and the fibrous cap was detected using dynamic programming. Our method has great potential to enable highly automated, objective, repeatable, and comprehensive evaluations of vulnerable plaques and treatments. We believe that the method is promising for both research and clinical applications.

## Data availability

The datasets generated and/or analyzed during the current study are not publicly available due to legal/ethical reasons but are available from the corresponding author on reasonable request.

## Acknowledgements

This project was supported by the National Heart, Lung, and Blood Institute through grants NIH R21 HL108263, NIH R01 HL114406, NIH R01 HL143484, and NIH R01 RES219220. This work was also supported by American Heart Association Grant #20POST35210974/Juhwan Lee/2020. This research was conducted in space renovated using funds from an NIH construction grant (C06 RR12463) awarded to Case Western Reserve University. The content of this report is solely the responsibility of the authors and does not necessarily represent the official views of the National Institutes of Health. The grants were obtained via collaboration between Case Western Reserve University and University Hospitals of Cleveland. This work made use of the High-Performance Computing Resource in the Core Facility for Advanced Research Computing at Case Western Reserve University. The veracity guarantor, Yazan Gharaibeh, affirms to the best of his knowledge that all aspects of this paper are accurate.

## Author contributions
J.L. was a major contributor in developing the methods and writing the manuscript. G.T.R.P. participated in the initial labeling and helped the comparison. Y.G. and C.K. helped to draft the manuscript and to analyze results. V.N.Z. and L.A.P.D. helped to perform the statistical analysis. J.N.K. and A.H. contributed to draft the manuscript. S.G.A.-K., G.G., and H.G.B. helped to design and validate the methods. D.L.W. supervised the research and helped to draft the manuscript. All authors read and approved the final manuscript.

## Competing interests
The authors declare no competing interests.

## Additional information
**Supplementary Information** The online version contains supplementary material available at https://doi.org/10.1038/s41598-022-24884-1.

**Correspondence** and requests for materials should be addressed to D.L.W.

**Reprints and permissions information** is available at www.nature.com/reprints.

**Publisher's note** Springer Nature remains neutral with regard to jurisdictional claims in published maps and institutional affiliations.